\documentclass[11pt,a4paper]{article}
\usepackage[hyperref]{acl2019}
\usepackage{times}
\usepackage{latexsym}
\usepackage{CJKutf8}
\usepackage{graphicx}
\usepackage{booktabs}
\usepackage{multirow}
\usepackage{amsmath}
\usepackage{bm}
\usepackage{url}
\usepackage{tabularx}
\usepackage{verbatim}
\usepackage{caption}
\usepackage{amsfonts}

\newcommand{\tabincell}[2]{\begin{tabular}{@{}#1@{}}#2\end{tabular}}

\aclfinalcopy 
\def\aclpaperid{1209} 

\title{ChID: A Large-scale Chinese IDiom Dataset for Cloze Test}

\author{Chujie Zheng$^{1,2}$, Minlie Huang$^1$\thanks{*Corresponding author: Minlie Huang.}, Aixin Sun$^3$ \\
  $^1$Institute for Artificial Intelligence, 
  Dept. of Computer Science, Tsinghua University\\
    $^1$Beijing National Research Center for Information Science and Technology, China \\
  $^2$Dept. of Physics, Tsinghua University, Beijing 100084, China \\
  $^3$School of Computer Science and Engineering, Nanyang Technological University, Singapore\\
  {\tt zcj16@mails.tsinghua.edu.cn, aihuang@tsinghua.edu.cn}\\{\tt axsun@ntu.edu.sg} \\
}

\date{}

\begin{document}

\newcolumntype{L}[1]{>{\raggedright\arraybackslash}p{#1}}
\newcolumntype{C}[1]{>{\centering\arraybackslash}p{#1}}
\newcolumntype{R}[1]{>{\raggedleft\arraybackslash}p{#1}}

\begin{CJK*}{UTF8}{gbsn}
\maketitle
\begin{abstract}
    Cloze-style reading comprehension in Chinese is still limited due to the lack of various corpora. In this paper we propose a large-scale Chinese cloze test dataset \textbf{ChID}, which studies the comprehension of idiom, a unique language phenomenon in Chinese. 
    In this corpus, the idioms in a passage are replaced by blank symbols and the correct answer needs to be chosen from well-designed candidate idioms. We carefully study how the design of candidate idioms and the representation of idioms affect the performance of state-of-the-art models. Results show that the machine accuracy is substantially worse than that of human, indicating a large space for further research.
  
\end{abstract}

\section{Introduction}

Machine reading comprehension aims to assess the ability to comprehend natural language and answer questions from a given document or passage. As a classical method of assessing language proficiency \cite{fotos1991cloze,jonz1991cloze,tremblay2011proficiency}, cloze test \cite{taylor1953cloze} has been widely employed due to its simplicity in form. Recently, a number of datasets for cloze test have been proposed for different languages. For instance, CNN/Daily Mail \cite{hermann2015teaching} provides a benchmark for machine comprehension of English text, while the People Daily and Children's Fairy Tale dataset \cite{cui2016consensus} and CMRC-2017 \cite{cui2017dataset} pioneer explorations in Chinese language. 

\begin{table}[t]
  \centering
  \scalebox{0.9}{
    \begin{tabular}{cc}
    \toprule
    \textbf{Idiom} & 亡羊补牢\\
    \midrule
    \textbf{Literal} & \tabincell{c}{To mend the fence after sheep are lost}  \\
    \cmidrule{1-2}
    \textbf{Meaning} & Never be late to try\\
    \bottomrule
    \end{tabular}
    }
    \caption{An example of metaphor in idiom. The sense of ``亡羊补牢'' should be inferred figuratively but not represented literally using the meanings of the four constituent characters.}
  \label{metaphor}
\end{table}

\begin{table}[t]
  \centering
  \scalebox{0.9}{
    \begin{tabular}{ccc}
    \toprule
    \textbf{Idioms} & 侃侃而谈 & 口若悬河 \\
    \midrule
    \textbf{Meanings} & \tabincell{c}{
    To speak with fervour\\and assurance} & \tabincell{c}{Fluently and\\eloquently} \\
    \midrule
    \textbf{Common} & \multicolumn{2}{c}{Speak much and long} \\
    \cmidrule{1-3}
    \tabincell{c}{\textbf{Difference}} & \tabincell{c}{Describe the deme-\\anor of the speaker} & \tabincell{c}{Describe the\\eloquence} \\
    \bottomrule
    \end{tabular}
    }
    \caption{An example of near-synonyms in idiom, where idioms share similar meanings but are different in language usage.}
  \label{near-synonyms}
\end{table}

\begin{table*}[t]
  \centering
  \scalebox{0.95}{
    \begin{tabular}{lcccllr}
    \toprule
    \textbf{Dataset} & \textbf{Lang.} & \textbf{Extractive} & \textbf{Option} & \textbf{Answer Type} & \textbf{Domain} & \textbf{Size}\\
    \midrule
    \tabincell{l}{CNN/Daily Mail
    } & EN & Yes & No & \tabincell{l}{Named entities} & News & 1.38M \\
    \tabincell{l}{Children's Book Test
    }   & EN & \tabincell{l}{Yes} & Yes & \tabincell{l}{Multiple types} & Children's Books & 678K \\
    \tabincell{l}{Who-did-What
    } & EN & \tabincell{l}{Yes} & Yes & \tabincell{l}{Named entities} & News & 206K \\
    \tabincell{l}{LAMBADA
    } & EN & \tabincell{l}{Partial}  & No & \tabincell{l}{Multiple types} & Novels &  10K \\
    \tabincell{l}{Story Cloze Test
    } & EN & No & Yes & Single sentence & Life Stories & 102K \\
    \tabincell{l}{CLOTH
    } & EN & No & Yes & \tabincell{l}{Multiple types} & Examinations & 99K \\
    \midrule
    \tabincell{l}{People Daily \& \\
    Children's Fairy Tale
    } & CN & Yes & No & \tabincell{l}{Nouns,\\ Named entities} & \tabincell{l}{News,\\Children's Stories} &  880K\\
    \tabincell{l}{CMRC-2017
    } & CN& Yes & No & \tabincell{l}{Nouns,\\Named entities} & Children's Stories & 359K\\
    \midrule
    \textbf{ChID (this work)} & \textbf{CN} & \textbf{No} & \textbf{Yes} & \tabincell{l}{\textbf{Chinese idioms}} & \tabincell{l}{\textbf{News, Novels,}\\\textbf{Essays}} & \textbf{729K}\\
    \bottomrule
    \end{tabular}
    }
    \caption{Comparison of ChID with other cloze-style reading comprehension datasets. \textit{Extractive} denotes whether the answer is extracted directly from the given context. \textit{Option} denotes whether candidate choices are provided. In the \textit{Answer Type} column, the answers of all datasets except Story Cloze Test are single words. \textit{Size} denotes the total number of queries or blanks in the dataset.}
  \label{comparison}
\end{table*}

In this paper we explore idiom comprehension
\cite{wray2002formulaic,jackendoff2002foundations,cacciari2014idioms,jiang2018chengyu} 
 in cloze test. Idiom
, which is called ``成语'' (chengyu) in Chinese, 
is an interesting linguistic phenomena in Chinese language, and this work is in parallel to several datasets \cite{Hill2015TheGP,xie2018large} that have considered different language phenomena in English.
Compared to other types of words, many idioms are unique for their non-compositionality and metaphorical meaning (see an example in Table \ref{metaphor}). This feature requires a good representation of idiom. Meanwhile, the characteristic of near-synonym, i.e., words that have similar but not identical meanings (see an example in Table \ref{near-synonyms}), may challenge a machine to choose an accurate idiom in a given context. Due to the fact that idioms are widely used in daily communication and in various literary genres, it is a new challenge to assess the ability of understanding and representing idioms in Chinese reading comprehension.

To this end, we propose \textbf{ChID}, a large-scale \textbf{Ch}inese \textbf{ID}iom dataset for cloze test. ChID contains 581K passages and 729K blanks, and covers multiple domains.
In ChID, the idioms in a passage were replaced with blank symbols. 
For each blank, a list of candidate idioms including the golden idiom are provided as choice. As the difficulty level of cloze test depends on candidate choices, we investigate several strategies of selecting candidate idioms. We evaluate 
several state-of-the-art models on the proposed corpus with different representations of idioms. Results show that machine performs much worse than human, which indicates a large room for further research.

Our contributions are summarized as follows:
\begin{itemize}
    \item We propose a new dataset, ChID, for cloze-style reading comprehension in Chinese language. ChID contains 581K passages and 729K blanks from three domains (news, novels, and essays). 
    
    \item We conduct extensive experiments on the design of candidate idioms and the idiom representation methods, and compare state-of-the-art models. Results show that the performance of these models is substantially worse than that of human. 
    
    \item ChID provides a benchmark to evaluate the ability of understanding idioms, a unique yet common language phenomenon in Chinese. To our knowledge, this is the first work where this linguistic phenomenon is studied in the form of machine reading comprehension.

\end{itemize}

\section{Related Work}


Recently, machine reading comprehension has been advanced by many corpora with various task settings. 
For instance, CNN/Daily Mail \cite{hermann2015teaching} collects news articles and uses the cloze test \cite{taylor1953cloze} to assess the ability of reading comprehension in English. RACE \cite{lai2017race} and CLOTH \cite{xie2018large} are constructed from questions in examinations designed for secondary and high school students. A number of question-answer datasets \cite{rajpurkar2016squad,reddy2018coqa} are also proposed and there are many other large-scale datasets \cite{nguyen2016ms,he2018dureader}. These corpora inspire various neural models \cite{chen2016thorough,cui2016consensus,seo2016bidirectional,dhingra2017gated,cui2017attention}.
In Table \ref{comparison}, we present a survey on existing cloze-style reading comprehension datasets. 

As the earliest cloze-style dataset for machine reading comprehension, CNN/Daily Mail \cite{hermann2015teaching} has a very large scale. 
It collects news articles paired with a number of bullet points, which summarise key aspects of an article. Based on the fact that these summary points are abstractive and do not simply copy sentences from a news article, 
the corpus is constructed by transforming these bullet points into cloze-style questions, i.e., replacing one entity with a placeholder. 
Children's Book Test (CBT) \cite{Hill2015TheGP} also provides a benchmark for machine reading comprehension, while the key differences from CNN/Daily Mail include: a list of candidate choices is provided for each query, and more types of words are removed, including named entities, (common) nouns, verbs and prepositions. 
Who-did-What \cite{onishi2016did} collects its corpus from news and provides options for questions similar to CBT. Each question is formed from two independent articles: an article is treated as context to be read and a separate article on the same event is used to form the query. 
LAMBADA \cite{paperno2016lambada} removed the last word from a given passage and evaluates the ability of word prediction.
By contrast, the Story Cloze Test dataset \cite{mostafazadeh2017lsdsem} evaluates the ability of story understanding and script learning, where the task requires to select or generate a reasonable sentence to complete the story context.

To the best of our knowledge, People Daily (PD) and Children's Fairy Tale (CFT) \cite{cui2016consensus} and CMRC-2017 \cite{cui2017dataset} are the only two existing cloze-style datasets for Chinese reading comprehension. Similar to CNN/Daily Mail and CBT, PD \& CFT and CMRC-2017 replaced a word (usually a noun or named entity) in the document with a blank placeholder and treated the sentence containing this word as a query. PD collects data from news while CFT and CMRC-2017 are from children's reading materials.


In most datasets, the answer can be directly found from context. CLOTH \cite{xie2018large} has a similar setting to ChID, where the answer should be selected from given choices. However, CLOTH is collected from English examinations for secondary/high school students, whose size is limited because documents, blanks, and options are all manually created. 

\section{Chinese Idioms}

Idiom is a common language phenomenon and usually called ``成语'' (chengyu) in Chinese. Thanks to its conciseness in form and expressiveness in meaning, idiom is widely used in daily communication and in various text genres. 
The main challenges for machine reading comprehension with idiom lie in: idiom representation which represents the meaning of an idiom, and thorough discrimination among the near-synonyms of an idiom.


\subsection{Idiom Representation}

Many idioms are non-compositional and have metaphorical meanings (see an example in Table \ref{metaphor}), which has also made idiom translation a challenging problem and attracted considerable research attentions \cite{anastasiou2010idiom,Salton2014EvaluationOA,cap2015account,Shao2018EvaluatingMT}. The meaning of such idioms is generally different from the literal meanings of the constituent characters. Such idioms are usually originated from ancient cultural stories, but the meaning is reserved along the long history of language use. For instance, ``塞翁失马'' has a metaphorical meaning, which is derived from this story: 
\begin{quote}
Near China's northern borders lived an old man who bred many horses. One day, one of his horses, for no reason at all, escaped into the territory of the northern tribes. Everyone commiserated with him. ``Perhaps this will soon turn out to be a blessing,'' said the old man. After a few months, his horse came back, and brought back a fine horse from the north.
\end{quote}
So the idiom ``塞翁失马'' usually refers to a blessing in disguise. 
Thus comprehending and representing an idiom may require the access to the corresponding cultural history. 
In addition, due to the polysemy of a single character, even those compositional idioms are likely to have ambiguity, which also makes idiom representation a challenging problem.

\subsection{Near-synonyms}

\begin{table*}[t]
  \centering
  \scalebox{1}{
    \begin{tabular}{ccrl}
    \toprule
    \multicolumn{2}{c}{\textbf{Passage \& Blanks}} &\multicolumn{2}{c}{ \tabincell{c}{
    可是有一个时期大家{\textit{\#idiom-0\#}}，不大敢露面，\\只有她一个人倚在阳台上看排队的兵走过。
    \\
    However, there was a period when everyone {\textit{\#idiom-0\#}}\\and was scared to show up. Only she leaned on the\\balcony and watched the soldiers passing by.
    }   }\\
    \midrule
    \multirow{8}{*}{ \tabincell{c}{\textbf{\textit{\#idiom-0\#}} \\\textbf{options}} } & \textbf{Correct} & 深居简出 & 
    Be unwilling to contact people 
    \\
    \cmidrule{2-4}
     & \textbf{Similar} & \quad \tabincell{r}{销声匿迹\\离群索居\\安分守己} & \tabincell{l}{
     To disappear from the scene
     \\
     To stay away from the crowd and live alone 
     \\
     To know one's place
     } \\
    \cmidrule{2-4}
     & \textbf{Random} & \tabincell{r}{一帆风顺\\文不对题\\万里长征} & \tabincell{l}{
     To proceed smoothly without a hitch
     \\
     Be irrelevant to the subject
     \\
     A long and arduous undertaking
     } \\ 
    \bottomrule
    \end{tabular}%
    }
    \captionof{figure}{An example in ChID. Each data contains a given passage with several blanks that replace the original idioms (in this example, there is only one blank). For each blank, several options are provided. Among the list of candidate choices, there is one golden answer, three similar idioms and another three random ones.}
  \label{data}
\end{table*}


It is common that an idiom has near-synonyms. These idioms may be confused in language use due to their similar but not identical meanings\footnote{Idioms with identical meanings can be interchangeably used in any context.} (see an example in Table \ref{near-synonyms}). To discriminate those near-synonyms, machine is required to figure out their subtle differences in usage, which is also challenging.

To verify the near-synonym phenomena, 
we conducted a user study. Based on the idiom vocabulary we collected (see Section 4.1), we manually evaluated the number of near-synonyms per idiom. We randomly sampled 200 idioms. For each idiom, we picked up the 20 most similar idioms whose embedding similarity score to the input idiom is less than some threshold. According to the similarity annotation result of Section 4.3 and Table \ref{similarity}, we set this threshold to 0.85. Then we hired four annotators to label these 4,000 idiom pairs in terms of whether a pair is near-synonyms or not. All the annotators have good command of Chinese.  

The evaluation result is shown in Table \ref{estimate}. Note that for each idiom, we rounded down the mean of the numbers of near-synonyms labeled by the four annotators. 
We estimate that about 90\% idioms have at least 1 near-synonym. About 23\% of the idioms have 4 or more near-synonyms. Fleiss' kappa \cite{fleiss1971measuring} for measuring inter-annotator agreement is 0.479, indicating moderate agreement (within [0.4, 0.6]).
This evaluation result strongly supports our claim that near-synonyms are very common among Chinese idioms.

\begin{table}[t]
  \centering
  \scalebox{1}{
        \begin{tabular}{ccc}
        \toprule
        $\ge \bm{K}$ \textbf{NEARs} & \# \textbf{Idioms} & \textbf{Proportion} \\
        \midrule
        1 & 179 & 89.5\%  \\
        2 & 131 & 65.5\% \\
        3 & 80 & 40.0\% \\
        4 & 40 & 23.0\% \\
        \bottomrule
        \end{tabular}%
    }
\caption{Annotation result of near-synonyms. It shows the number of idioms in the 200 sampled idioms that have at least $K$ near-synonyms, for $K=1,2,3,4$. Fleiss' kappa is 0.479, indicating moderate agreement.}
  \label{estimate}%
\end{table}

\section{Dataset Collection}

Figure \ref{data} presents an example in ChID. In each sample, idioms in a passage are replaced by blank symbols, and each blank is provided with several candidate idioms including the golden idiom. The task is to select the golden answer from the candidate choices given the context. Note that the answer is usually not occurring in the context in our setting, which is different from most existing cloze test corpora. 

In the following subsections, we will explain the three steps in data collection: (1) Constructing the idiom vocabulary; (2) Extracting passages within a proper length; (3) Designing candidate choices.


\begin{table}[t]
  \centering
    \scalebox{1}{
        \begin{tabular}{clrr}
        \toprule
        \textbf{Level} & \textbf{Freq.} & \multicolumn{1}{c}{\textbf{Num.}} & \multicolumn{1}{c}{\textbf{Prop.}} \\
        \midrule
        \textbf{Very Low} & {[20, 50)} &  832 & 21.6\% \\
        \textbf{Low} & {[50, 100)} &  742 & 19.3\% \\
        \textbf{Medium} & {[100, 200)} &  822 & 21.4\% \\
        \textbf{High} & {[200, 400)} &  746 & 19.4\% \\
        \textbf{Very High} & {[400, 534]} & 706 & 18.3\% \\
        \cmidrule{1-4}
        \textbf{Total} & [20, 534] & 3,848 & 100.0\% \\ 
        \bottomrule
        \end{tabular}%
    }
    \caption{Idiom frequency statistics in the whole corpus. The minimum and the maximum are 20 and 534 respectively.}
  \label{frequency}%
\end{table}%

\subsection{Vocabulary Construction}

We collected the idiom vocabulary from Chinese idioms Daquan
\footnote{\tt http://www.guoxue.com/chengyu/CYML.htm}, which contains over 23K idiom entries. 
Since vast majority of idioms consist of 4 characters, we only retained idioms with 4 characters in our vocabulary. 
In order to facilitate the design of candidate choices, we removed those idioms that do not have a pre-trained embedding using the large-scale open-source corpus provided by \citet{song2018directional}, where approximately 40\% idioms were filtered out.
We normalized synonyms with only slight morphological variation. Idioms that share the same explanation and meaning, but only differ in one character or the order of characters, are treated as the same idiom. This can be done with the Chinese idiom dictionary because some idioms are marked with: ``又作'' (also written as), ``犹'' (like), ``同'' (the same as), ``见'' (also see).
Such idioms in the passages are all replaced by their re-normalized ones. 

We then counted the frequency of each idiom in the corpus, and removed those idioms that appear less than 20 times. Finally, the idiom vocabulary has 3,848 entries in total, and their frequency statistics on the whole corpus is shown in Table \ref{frequency}. The minimum and the maximum idiom frequencies are 20 and 534 respectively. We simply divide the idiom frequency into five intervals: very low (from 20 to 50), low (from 50 to 100), medium (from 100 to 200), high (from 200 to 400) and very high (higher than 400). The proportions of idioms in the frequency intervals are almost uniformly distributed.


\subsection{Passage Extraction}
To make the topic and domain more diversified, we collected passages from novel and essay on the Internet, and the news articles provided by \citet{sun2016thuctc}\footnote{The news articles are extracted from the THUCNews dataset for Chinese text classification, which can be downloaded from {\tt http://thuctc.thunlp.org/}.}. Since some documents may be very long, we took a paragraph as the basic unit. Each idiom except those in double quotation marks\footnote{Because words in a quotation mark are usually entities or other content that can not be inferred from the context.} 
is replaced with a blank symbol. A paragraph that is shorter than 100 characters is merged with the next paragraph to ensure that the context are sufficient for answer selection. Those passages that are longer than 600 characters are abandoned.  

It is worth noting that if some idiom has a much higher word frequency than others, models may tend to bias answer selection to those more frequent idioms. In order to make frequent and infrequent idioms more balanced, we removed some passages which only contain high frequency idioms.

\subsection{Candidate Choice Selection}

The semantic relevance between two idioms can be measured by the cosine similarity of their embeddings \cite{Mikolov2013EfficientEO}, which helps us to design candidate choices. However, idioms that are similar in embedding may or may not be synonyms or near-synonyms. We thus manually evaluated the correlation between embedding similarity and idiom synonymity.
We split the embedding similarity from 0.9 to 0.5 into 8 intervals. Within each interval, 200 pairs of idioms are sampled. We used three labels to measure the relevance between two idioms: \textbf{SYN} (synonym, the two idioms are identical in meaning and can be interchangeably used), \textbf{NEAR} (near-synonym, have close or similar meanings but can not be used interchangeably), \textbf{OTHER} (irrelevant or opposite in meaning). We hired five annotators to label these samples.

\begin{table}[t]
  \centering
  \scalebox{0.9}{
        \begin{tabular}{lrrrc}
        \toprule
        \textbf{Similarity} & \textbf{SYN} & \textbf{NEAR} & \textbf{OTHER} & $\bm{\kappa}$ \\
        \midrule
        {[0.85, 0.90)} & \textbf{83.2\%} & 16.8\% & 0.0\% &  .642 \\
        {[0.80, 0.85)} & \textbf{53.6\%} & 42.8\% & 3.6\% &  .447 \\
        {[0.75, 0.80)} & 29.2\% & \textbf{53.6\%} & 17.2\% &  .485 \\
        {[0.70, 0.75)} & 12.0\% & \textbf{57.2\%} & 30.8\% &  .496 \\
        {[0.65, 0.70)} & 0.4\% & \textbf{52.8\%} & 46.8\% &  .466 \\
        {[0.60, 0.65)} & 0.0\% & 34.0\% & \textbf{66.0\%} &  .528 \\
        {[0.55, 0.60)} & 0.0\% & 10.4\% & \textbf{89.6\%} &  .657 \\
        {[0.50, 0.55)} & 0.0\% & 6.0\% & \textbf{94.0\%} &  .787 \\
        \bottomrule
        \end{tabular}%
    }
\caption{Annotation result of embedding similarity. The three labels are: \textbf{SYN} (synonym), \textbf{NEAR} (near-synonym), \textbf{OTHER}. $\bm{\kappa}$ is the Fleiss' kappa value. }
  \label{similarity}%
\end{table}

\begin{table*}[t]
  \centering
  \scalebox{1}{
    \begin{tabular}{c|rrrr|r|r}
    \toprule
    \multicolumn{1}{r|}{} & \multicolumn{4}{c|}{\textbf{In-domain}} & \multicolumn{1}{c|}{\textbf{Out-of-domain}} & \multirow{2}{*}{\textbf{Total}} \\
    \cmidrule{2-6}
     & \textbf{Train} & \textbf{Dev} & \textbf{Test} & \textbf{Total} & \textbf{Out}  \\
    \midrule
    \textbf{\# Passages} & 520,711 & 20,000 & 20,000 & 560,711 & 20,096 & 580,807  \\
    \textbf{Avg. \# tokens per passage} & 99 & 99 & 99 & 99 & 127 & 100  \\
    \midrule
    \textbf{\# Distinct idioms covered} & 3,848 & 3,458 & 3,502 & 3,848 & 3,626 & 3,848 \\
    \textbf{Avg. idiom frequency} & 168.6 & 7.2 & 7.1 & 181.6 & 8.3 & 189.6 \\
    \midrule
    \textbf{Total \# blanks} & 648,920 & 24,822 & 24,948 & 698,690 & 30,023 & 728,713 \\
    \textbf{Avg. \# blanks per passage} & 1.25 & 1.24 & 1.25 & 1.25 & 1.49 & 1.25 \\
    \textbf{Single-blank prop.} & 80.4\% & 80.7\% & 80.8\% & 80.5\% & 64.7\% & 79.9\% \\
    \textbf{Multi-blank prop.} & 19.6\% & 19.3\% & 19.2\% & 19.5\% & 35.3\% & 20.1\% \\
    \bottomrule
    \end{tabular}}
    \caption{ChID dataset statistics. The out-of-domain data have longer passages (127 vs. 99) and more blanks per passage (1.49 vs. 1.25) than the in-domain data. 
    }
  \label{statistics}
\end{table*}

As shown in Table \ref{similarity}, when the similarity score is larger than 0.75, there is a large proportion of idioms pairs that have the same meaning; when the score is between 0.65 and 0.80, there is a large probability that the two idioms are near-synonyms. For those pairs with high (larger than 0.85) or low (smaller than 0.60) similarity, annotators tend to reach substantial agreement\footnote{Substantial agreement corresponds to kappa within [0.6, 0.8].} according to Fleiss' kappa, while we have moderate agreement between the similarity interval [0.65, 0.85].

The above annotation results inspire us to design proper candidate choices for each blank in a passage. First of all, we excluded those idioms that have a similarity score higher than 0.7 to the golden answer. This avoids to include synonyms of the golden answer in the candidate choice. 
Then, we picked up top 10 similar idioms among the remaining idioms, and randomly chose three idioms as candidate choice. 
Note that the three idioms have a large probability of being near-synonyms of the golden answer, which affects the difficulty level of the cloze test to some degree.
We further randomly sampled another three idioms from the remaining idioms that do not include the top 10 similar idioms. 
In this manner, the list of candidate choices consists of three parts: the correct answer, three similar idioms, and three other randomly sampled ones, as shown in Figure \ref{data}.

\subsection{Corpus Statistics}

\begin{table}[t]
  \centering
    \scalebox{1}{
        \begin{tabular}{clrr}
        \toprule
        \textbf{Level} & \textbf{Freq.} & \multicolumn{1}{c}{\textbf{In}} & \multicolumn{1}{c}{\textbf{Out}} \\
        \midrule
        \textbf{Very Low} & {[20, 50)} &  3.5\% & \textbf{8.2\%} \\
        \textbf{Low} & {[50, 100)} &  7.2\% & \textbf{12.0\%} \\
        \textbf{Medium} & {[100, 200)} &  16.0\% & 19.7\% \\
        \textbf{High} & {[200, 400)} &  28.8\% & 28.7\% \\
        \textbf{Very High} & {[400, 534]} & 44.5\% & \textbf{31.4\%} \\
        \cmidrule{1-4}
        \textbf{Total} & [20, 534] & 100.0\% & 100.0\% \\ 
        \bottomrule
        \end{tabular}%
    }
    \caption{Comparison on idiom frequency distribution between the in-domain and out-of-domain data. 
    }
  \label{inout}%
\end{table}%

The detailed statistics of ChID is shown in Table \ref{statistics}. News and novels are treated as in-domain data, which are divided into the training set \textbf{Train}, the development set \textbf{Dev}, and the test set \textbf{Test}. Essays are reserved for out-of-domain test \textbf{Out} to assess the generalization ability of cloze test models. The in-domain data cover 3,848 Chinese idioms,  while \textbf{Dev}/\textbf{Test}/\textbf{Out} respectively cover 3,458/3,502/3,626 idioms.

There are some differences between in-domain and out-of-domain data. 
Firstly, the average length of passages in the in-domain data is nearly 100 words, while Out-of-domain data have longer passages (127 words). The average number of blanks per passage is also different (1.25 vs. 1.49). 
Secondly, the idiom distributions are different. As shown in Table \ref{inout}, compared to the in-domain data, low-frequency idioms occupy a higher proportion of all the idiom occurrences in the out-of-domain data (8.2\% vs. 3.5\% for very low frequency interval and 12.0\% vs. 7.2\% for low frequency interval) while the high-frequency idioms occur less frequently (31.4\% vs. 44.5\%). These differences make the out-of-domain test set more challenging.

\section{Experiment}


\subsection{Models}
In order to evaluate how well the state-of-the-art models can comprehend Chinese language with idiom, we tested the following models:

\noindent \textbf{Language Model} (LM): 
We trained a bidirectional LSTM \cite{hochreiter1997long} to obtain the hidden state at the blank ($\bm{h}_b$), and use the hidden state to score candidate choices:
\begin{align}
    \overrightarrow{\bm{h}}_b = \overrightarrow{\textbf{LSTM}}(&w_{1:b}),\ \overleftarrow{\bm{h}}_b = \overleftarrow{\textbf{LSTM}}(w_{b:|\mathbf{p}|})\\
    \bm{h}_b&=\overrightarrow{\bm{h}}_b\oplus \overleftarrow{\bm{h}}_b\\
    \alpha_i&=\mathrm{softmax}_i\left(\bm{h}_b^T\bm{c}_i\right)
\end{align}
where $|\bf{p}|$ denotes the length of passage $\bf{p}$, 
$w_{1:b},w_{b:|\mathbf{p}|}$ denote the words in the given context before or after the blank respectively, 
$\oplus$ denotes concatenation, and $\bm{c}_i$ denotes the embedding of each candidate idiom. Then, the option that has the highest $\alpha_i$ is chosen as the answer.

\noindent \textbf{Attentive Reader} (AR) \cite{hermann2015teaching}: The bidirectional LSTM model is augmented with the attention mechanism \cite{bahdanau2014neural}. The hidden state at blank $\bm{h}_b$ is used as the query to attentively read the context as follows: 
\begin{align}
    \bm{m}_t&=\mathrm{tanh}(\bm{W}_{hm}\bm{h}_t+\bm{W}_{bm}\bm{h}_b)\\
    s_t&=\mathrm{softmax}_t\left(\bm{w}^T_{ms}\bm{m}_t\right),\ \bm{r}=\sum_{t=1}^{|\mathbf{p}|}s_t\bm{h}_t
\end{align}
where $\bm{W}_{hm}, \bm{W}_{bm}, \bm{w}_{ms}$ are all parameters. Then, the attention vector $\bm{r}$ and the blank vector $\bm{h}_b$ are used to score each candidate choice:
\begin{align}
    \bm{g}&=\mathrm{tanh}(\bm{W}_{rg} \bm{r} + \bm{W}_{bg} \bm{h}_b)\\
    \alpha_i&=\mathrm{softmax}_i\left(\bm{g}^T\bm{c}_i\right)
\end{align}
where $\bm{W}_{rg}, \bm{W}_{bg}$ are also parameters. 

\noindent \textbf{Stanford Attentive Reader} (SAR) \cite{chen2016thorough}: Compared to AR, SAR applies a bilinear matrix $\bm{W}_s$ to compute attention weights instead of using a tanh layer. The weighted contextual vector $\bm{o}$ is used for scoring candidates:
\begin{align}
    s_t &= \mathrm{softmax}_t\ \left(\bm{h}_b^T\bm{W}_s\bm{h}_t\right),\ \bm{o}=\sum_{t=1}^{|\mathbf{p}|} s_t\bm{h}_t\\
    \alpha_i&=\mathrm{softmax}_i\left(\bm{o}^T\bm{c}_i\right)
\end{align}

\subsection{Implementation Details}
All the models were implemented with Tensorflow \cite{abadi2016tensorflow}. We employed the Jieba Chinese word segmenter\footnote{\tt https://github.com/fxsjy/jieba} to tokenize passages. We set the vocabulary size to 100K and used the 200-dimensional word embeddings initialized by \citet{song2018directional}. Those word embeddings that were not matched in \citet{song2018directional} were initialized from a uniform distribution between (-0.1, 0.1).
We applied a dropout rate of 0.5 on word embeddings. The number of hidden units of RNN cells were all set to 100. The cross entropy cost function is used to compute the training loss. ADAM \cite{kingma2015adam} was used to optimize all the models with the initial learning rate to 0.001 and the gradient was clipped when the norm of the gradient was larger than 5. We set the batch size to 32. The training was stopped when the accuracy on \textbf{Dev} did not improve within an epoch.


\subsection{Option Settings}

To evaluate how the method of candidate choice design will impact the performance, we prepared two additional test sets:  \textbf{Ran} and \textbf{Sim}, both of which have the same passages with \textbf{Test}, but candidate choices are designed differently. 
In \textbf{Ran}, all the candidate choices are sampled from the idioms that are not similar to the golden answer. Instead, in \textbf{Sim}, all the candidates are sampled from top 10 similar idioms. Therefore, \textbf{Sim} is more challenging than \textbf{Ran} as the former has more distracting options. Note that each blank has seven choices including the golden answer.

\begin{table}[t]
  \centering
    \scalebox{1}{
        \begin{tabular}{lccccc}
        \toprule
        & \textbf{Dev} & \textbf{Test} & \textbf{Ran} & \textbf{Sim} & \textbf{Out} \\
        \midrule
        \textbf{Human} & - & \textbf{87.1} & \textbf{97.6} & \textbf{82.2} & \textbf{86.2} \\
        {$\bm{\kappa}$} & {-} & {.794} & {.953} & {.791} & {.769} \\
        \cmidrule{1-6}
        \textbf{LM} & 71.8 & 71.5 & 80.7 & 65.6 & 61.5 \\
        \textbf{AR} & \textbf{72.7} & \textbf{72.4}  & \textbf{82.0} & \textbf{66.2} & \textbf{62.9} \\
        \textbf{SAR} & 71.7 & 71.5 & 80.0 & 64.9 & 61.7 \\
        \bottomrule
        \end{tabular}%
    }
    \caption{Performance of human and models. $\bm{\kappa}$ indicates Fleiss' kappa. The overall best results are shown in bold, and AR performs significantly better than LM and SAR (sign test, $p$-value $<$ 0.05).}
  \label{result}%
\end{table}%

\subsection{Results}

To explore the ceiling of model performance, we also conducted  \textbf{Human Evaluation}. 
We sampled 200 passages 
respectively from the aforementioned test sets: \textbf{Test}, \textbf{Ran}, \textbf{Sim} and \textbf{Out}. We then hired three annotators to complete the 800 cloze tests. These three annotators are first-year or second-year university students and all have very good command of Chinese language. The average accuracy of the annotators and the corresponding Fleiss' kappa are reported as the final performance.

The experiment results are shown in Table \ref{result}. We analyzed the results from the 
following perspectives:

\textbf{Option Setting:}
The setting of similar options is much harder than that of random options. Firstly, we noted that both human and models achieve worse performance on \textbf{Test} than on \textbf{Ran}, while the accuracy on \textbf{Sim} is even lower than \textbf{Test}, which indicates that including more similar candidate idioms makes the task more difficult. Secondly, the
inter-annotator agreement on \textbf{Ran} (Fleiss' kappa=0.953) is much higher than those on other test sets which include similar options. This implies that similar options also make manual annotation harder.


\textbf{Human vs. Models:}
 Firstly, human performance is substantially better than model performance on all the test sets. The smallest gap between human and machine is 14.6 (on \textbf{Test}) and the largest gap is 23.3 (on \textbf{Out}). Secondly, humans perform very closely on \textbf{Test} and \textbf{Out} (87.1 vs. 86.2), 
however, the models perform much better on \textbf{Test} than on \textbf{Out} (72.4 vs. 62.9). This observation implies that human has a strong ability to generalize to out-of-domain data while the models cannot generalize well to \textbf{Out} which contains more low-frequency idioms. 

\textbf{Model Comparison:}
AR outperforms all other models significantly. The reason for this may be due to the fact: AR firstly uses the blank representation ($\bm{h}_b$) to make an attentive read of the context (see Eq. 4 and 5), and the blank vector is used again with the attentive vector ($\bm{r}$) to score a candidate choice. In this manner, the context is attentively used and the blank vector is used twice.

\subsection{Comparison on Idiom Representation}

\begin{table}[t]
  \centering
    \scalebox{1}{
        \begin{tabular}{lccccc}
        \toprule
        & \textbf{Dev} & \textbf{Test} & \textbf{Ran} & \textbf{Sim} & \textbf{Out} \\
        \midrule
        \midrule
        \multicolumn{6}{c}{\textbf{Idiom Embedding}} \\
        \midrule
        \textbf{LM} & 71.8 & 71.5 & 80.7 & 65.6 & 61.5 \\
        \textbf{AR} & 72.7 & 72.4  & 82.0 & 66.2 & 62.9 \\
        \textbf{SAR} & 71.7 & 71.5 & 80.0 & 64.9 & 61.7 \\
        \midrule
        \midrule
        \multicolumn{6}{c}{\textbf{Average Character Embedding}} \\
        \midrule
        \textbf{LM} & 63.1 & 63.0 & 70.6 & 57.2 & 53.2 \\
        \textbf{AR} & 64.5 & 63.8 & 72.5 & 57.8 & 53.5 \\
        \textbf{SAR} & 62.9 & 62.5 & 71.4 & 56.9 & 52.0 \\
        \midrule
        \midrule
        \multicolumn{6}{c}{\textbf{Average Character Embedding + MLP}} \\
        \midrule
        \textbf{LM} & 68.6 & 68.4 & 77.4 & 62.0 & 57.8 \\
        \textbf{AR} & 67.1 & 66.5 & 75.6 & 60.7 & 56.4  \\
        \textbf{SAR} & 67.0 & 66.8 & 75.6 & 60.8 & 55.6 \\
        \bottomrule
        \end{tabular}%
    }
    \caption{Performance comparison using different idiom representations. 
    }
  \label{com}%
\end{table}%

In previous experiments, an idiom was treated as a token, and its representation are obtained through pretraining on a large corpus \cite{song2018directional}. 
In this section, we explored another two methods for idiom representation, and evaluated the performance with different idiom representations. 
One method simply uses the average embedding of 4 constituent characters as the representation of an idiom. This method mimics to understand idioms purely based on its literal meanings. 
The other is to apply an MLP (Multi-Layer Perceptron, \citealp{bishop1995neural,fine1999feedforward}) which is fed with the concatenation of 4 character embeddings, and the output vector is used to represent an idiom. This method also applies a composition assumption: the representation of an idiom is a composite function of its constituent words. 
Note that the input to the MLP is an 800-dimension vector, and the MLP has a hidden layer of 400 units and uses tanh as the activation function. The final output of MLP is a 200-dimension vector. 

Table \ref{com} shows the performance comparison using three methods for idiom representation. We can observe remarkable drops from \textit{idiom embedding} to \textit{average character embedding + MLP} and to \textit{average character embedding} for all the models, where all the differences are significant (sign test, $p$-value $<0.01$). 
The results indicate that the other two idiom representation methods are worse than treating an idiom as an independent semantic unit. 
This study also implies that idiom representation is a key factor for the success of Chinese reading comprehension with idiom. In other words, a good cloze test model should have not only a proper model structure, but also a good method to represent idioms.

\section{Conclusion}



In this paper, we propose a large-scale Chinese cloze dataset (ChID) which contains 581K passages and 729K queries from news, novels, and essays, covering 3,848 Chinese idioms. The corpus provides a benchmark to evaluate the ability of Chinese cloze test with idiom. Firstly, we analyze how the embedding similarity correlates with synonymity and near-synonymity of Chinese idiom, and find that the difficulty level of Chinese cloze test with idiom correlates positively with the method of choosing candidate choices. 
Secondly, we find that idiom representation is a key factor to the success of reading comprehension models in this task due to the common non-compositionality and metaphorical meaning of Chinese idiom. 
Thirdly, we evaluate three state-of-the-art cloze test models on this corpus, and observe that existing model performance is still much worse than human performance. 
All these findings indicate that the corpus may be a proper benchmark for Chinese cloze test and worth further research\footnote{Our dataset is available at {\tt https://github.com/chujiezheng/ChID-Dataset}.}.


\section*{Acknowledgments}
This work was jointly supported by the National Science
Foundation of China (Grant No.61876096), and
the National Key R\&D Program of China (Grant No.
2018YFC0830200).

\bibliography{acl2019}
\bibliographystyle{acl_natbib}

\end{CJK*}
\end{document}